\documentclass[journal]{IEEEtran}
\usepackage{amsmath,amsfonts,amssymb}
\usepackage{algorithmic}
\usepackage{algorithm}
\usepackage{array}
\usepackage[caption=false,font=normalsize,labelfont=sf,textfont=sf]{subfig}
\usepackage{textcomp}
\usepackage{stfloats}
\usepackage{url}
\usepackage{verbatim}
\usepackage{graphicx}
\usepackage{cite}
\usepackage{tabularx}
\usepackage[table]{xcolor}
\begin{document}

\title{A Generative Model for Closed-Loop Microsimulation of Signalized Intersections}

\author{Yash~Ranjan,
        Rahul~Sengupta,
        Anand~Rangarajan,
        and~Sanjay~Ranka,~\IEEEmembership{Fellow,~IEEE}%
\thanks{The authors are with the Department of Computer and Information Science and Engineering, University of Florida, Gainesville, FL 32611, USA.
E-mail: \{yashranjan, rahulseng\}@ufl.edu, \{anand, ranka\}@cise.ufl.edu.}%
\thanks{A preliminary version of this work appeared in~\cite{enactor_vehits}.}}

\markboth{IEEE Transactions on Intelligent Transportation Systems}%
{Ranjan \MakeLowercase{\textit{et al.}}: A Generative Model for Closed-Loop Microsimulation of Signalized Intersections}

\maketitle

\begin{abstract}
Traffic microsimulators rely on hand-crafted behavior models that reproduce aggregate flow but miss the heterogeneous interactions between vehicles at signalized intersections. Learned trajectory predictors capture richer interactions but are short-horizon and tend to be unstable when run in closed loop. We present \emph{Enactor}, an actor-centric generative model for closed-loop intersection microsimulation. The model focuses on vehicles; pedestrians are included as context that can influence vehicle decisions but are not predicted directly. Dynamic actors and lane polylines are encoded in polar coordinates referenced to the intersection center, and each actor sees the rear-bumper position of its leader so that deceleration can be terminated at the right place. A transformer with separate spatial and temporal attention blocks predicts a distribution over each actor's next-step motion ($s$, $\alpha$); positions are propagated by a deterministic kinematic update. Training uses a closed-loop curriculum so the model is exposed to its own predictions before it has to rely on them at inference.

We evaluate Enactor in two regimes. In a $4{,}000$-second simulation-in-the-loop test at two intersection geometries, Enactor controls every dynamic vehicle against a continuously refreshing actor set rather than the fixed cohort that learned trajectory predictors are usually evaluated against. It recovers the SUMO data generator's speed and travel-time distributions with KL divergence over an order of magnitude lower than a recent transformer baseline on travel time, and substantially lower on speed (roughly $5\times$ lower at Site 1), and reduces red-light violations relative to the same baseline by more than an order of magnitude. An ablation isolates the leader rear-bumper feature as the change with the largest effect on intersection-aware safety metrics; close-range interaction realism (TTC) remains the main open problem. We also evaluate on real-world field data. We apply the same architecture to naturalistic vehicle trajectories from a fish-eye camera at a signalized intersection, with a multi-horizon predictive evaluation that fits the single-pass nature of the recording. Enactor outperforms a constant-velocity baseline at every horizon evaluated.
\end{abstract}


\section{Introduction}
\label{sec:intro}
\IEEEPARstart{T}{raffic} microsimulators such as SUMO and MATSim are widely used to evaluate road-network performance under what-if conditions: intersection design, signal timing, demand management. The behavior models that drive their actors are typically hand-crafted and calibrated at the population level. Car-following models like the Intelligent Driver Model and the Krauss model assume smooth, rational drivers; pedestrian dynamics rely on force-based formulations such as the Social Force Model. These models work well for aggregate flow but reproduce poorly the heterogeneous, sometimes aggressive or distracted behavior that occurs at intersections, and the data needed to calibrate actor-specific parameters is rarely available~\cite{NI2020102137}.

Data-driven trajectory predictors learn multi-agent interactions from observed trajectories~\cite{salzmann2021,shi2023mtr}, and generative variants are now used to simulate the motion of surrounding agents in autonomous-driving contexts~\cite{trafficbots2023}. Two problems remain. When such models are run in closed loop over long horizons, prediction errors compound and the rollout drifts away from the training distribution. And capturing physically plausible, socially consistent behavior even at short horizons usually requires large amounts of high-quality data, because the model has to learn physical constraints and interaction dynamics implicitly from observation alone.

Intersections are a tough case for both kinds of approach. They are conflict zones where vehicle and pedestrian trajectories cross, where a substantial fraction of crashes occur, and where stochastic driver decisions interact with strong structural constraints from geometry, lane topology, and signal timing. We exploit that structure. \emph{Enactor} is an actor-centric generative model that combines a polar-coordinate state representation aligned with intersection geometry, a transformer-based interaction module, and a closed-loop training schedule that exposes the model to its own predictions during training. Each actor also sees the rear-bumper position of its leader on the same lane, when one exists; this geometric cue tells the model where deceleration should end, and our ablations show it is the single change with the largest effect on intersection-aware safety metrics. We argue that stable long-horizon closed-loop rollout, aggregate-flow realism, and intersection-aware safety realism are three separate objectives, addressed by different design choices, and that doing well on one does not automatically translate into doing well on the others.

We evaluate Enactor in two regimes. On data generated by SUMO at two intersection geometries, we run a closed-loop simulation-in-the-loop protocol in which Enactor controls every dynamic actor for $40{,}000$ timesteps ($4{,}000$s). On naturalistic data from a fish-eye camera at a signalized intersection, processed through a YOLO-based detection-and-tracking pipeline, we evaluate single-pass per-actor prediction at multi-second horizons. The two regimes answer different questions because they have to: real-world recordings capture one realization of the world and offer no replayable ground-truth alternative future, so closed-loop rollout against ground truth is not available there.

\paragraph*{Scope: vehicles as targets, pedestrians as context} Pedestrians are present at signalized intersections and can influence vehicle behavior: a vehicle yields, decelerates, or accelerates differently depending on whether a pedestrian is in or near the conflict zone. We include pedestrians in the model as neighbors --- their state enters each vehicle's spatial-attention context and indirectly influences its predicted trajectory. We do not study pedestrian behavior itself: pedestrian trajectories are not evaluation targets, and we make no claim about the realism of pedestrian motion the model might generate. Treating pedestrians as first-class targets is left for future work.

A central feature of the simulator regime is that the cohort of dynamic actors changes over time: actors enter the intersection on one approach, traverse it, leave on another, while new actors arrive continuously at the perimeter. Most learned trajectory predictors evaluate against a fixed cohort of actors at fixed prompt times, which avoids the question of how the model behaves when the scene graph it has to reason over is itself non-stationary. Sustaining a coherent closed-loop rollout under a continuously refreshing actor set, without diverging from the road network or producing implausible trajectories over thousands of timesteps, is one of the main technical problems this paper addresses. Both Enactor and the IntTrajSim baseline are evaluated under the same $4{,}000$-second SUMO-in-the-loop protocol against this non-stationary actor cohort, and the resulting metrics measure relative rule compliance and aggregate-flow realism inside that protocol. Naturalistic trajectories bring a separate problem the simulator does not have: bounding-box jitter from object detection, especially perpendicular to the direction of motion, produces artificially large frame-to-frame velocities and accelerations, and we evaluate both smoothed and raw variants to see how much this upstream noise affects what the model learns.

The contributions of this paper are:
\begin{itemize}
    \item a polar-coordinate, geometry-aware state representation for intersection actors and map elements;
    \item a transformer-based generative model trained in closed loop and evaluated as a full microsimulator in SUMO across two intersection geometries, sustaining stable rollouts over $4{,}000$~seconds against a continuously refreshing actor cohort rather than a fixed-cohort evaluation;
    \item an ablation isolating the contribution of including the leader's rear-bumper position as an input feature;
    \item a direct comparison against IntTrajSim under the same closed-loop protocol, reducing red-light violations by more than an order of magnitude ($5$ vs.\ $169$ or $175$, depending on the baseline's post-processing) at the first intersection and showing similarly large gaps at the second, together with trajectory-reconstruction benchmarks where ADE/FDE are the primary cross-model comparison and NLL is reported under each model's own predictive distribution; and
    \item an evaluation on real-world field data, by applying the same architecture to naturalistic vehicle trajectories under a multi-horizon predictive evaluation, where Enactor outperforms a constant-velocity baseline at $2$, $5$, and $10$~second horizons.
\end{itemize}

The remainder of the paper is organized as follows. Section~\ref{sec:related} surveys related work in trajectory forecasting, scene-level world models, and intersection-specific simulation and evaluation. Section~\ref{sec:problem} formalizes the modeling task. Section~\ref{sec:arch} describes the Enactor architecture and the closed-loop training procedure. Section~\ref{sec:data} describes the simulator and real-world data sources together with the preprocessing applied to the latter. Sections~\ref{sec:exp_sim} and~\ref{sec:exp_real} report the simulator and real-world studies. Section~\ref{sec:conclusion} concludes.

\section{Related Works}
\label{sec:related}

We organize related work into four groups: foundational interaction-aware trajectory prediction, attention-based prediction architectures, prediction at signalized and unsignalized intersections, and scene-level world models for autonomous driving.

\textbf{Interaction-aware trajectory prediction.} A first generation of learned trajectory predictors modeled an actor's future motion as a function of its own history and a learned representation of nearby agents. Social LSTM~\cite{alahi2016sociallstm} introduced a pooling mechanism that lets recurrent encoders share hidden states across pedestrians in a crowd. Convolutional social pooling~\cite{deo2018convolutional} extended the same idea to vehicles on highways using a spatial grid. Social GAN~\cite{gupta2018socialgan} added a generative formulation, sampling diverse plausible futures rather than committing to a single mode. Trajectron++~\cite{salzmann2021} combined LSTM-based history encoding with graph attention for heterogeneous-agent interaction. These models are short-horizon predictors evaluated open-loop against ground truth; they are not designed to control the dynamics of all surrounding actors recursively over thousands of timesteps and they typically do not model traffic-signal state.

\textbf{Attention-based prediction architectures.} Since~\cite{vaswani2017}, attention has displaced recurrence as the default sequence-modeling mechanism for trajectory prediction, in part because it scales better with the number of interacting agents. \cite{messaoud2021attention} applied multi-head attention directly to neighboring vehicles for highway prediction. AI-TP~\cite{zhang2023aitp} formalized interaction-awareness through learned attention weights between an ego actor and its neighbors. \cite{chen2022intentionaware} factored the prediction into spatial and temporal attention blocks, a decomposition we also adopt. The hierarchical transformer of~\cite{geng2023transferable} addresses heterogeneous agent classes through transferable attention modules. MTR~\cite{shi2023mtr} achieves state-of-the-art motion forecasting through a global intention-localization module combined with local refinement. HDGT~\cite{hdgt2023} formulates the scene as a heterogeneous graph and applies graph transformers to encode it. The attention literature is the architectural lineage of Enactor; what differs in our work is not the use of attention but the closed-loop training and evaluation regime, the polar state representation, and the integration of leader-rear-bumper geometric cues directly into the actor's input.

\textbf{Prediction at signalized and unsignalized intersections.} A second body of work specializes trajectory prediction to the intersection setting, where signals and lane geometry constrain plausible behavior. Bayesian network models for red-light running~\cite{chen2019bayesian} and turning intention~\cite{zhang2021turning} encode signal-phase and lane-geometry priors explicitly. Hierarchical predictors~\cite{yang2023hierarchical} couple signal state to interactive agent behavior in a structured way. \cite{oh2019impact} quantifies the effect of including signal phase as a forecasting input near signalized intersections, and D2-tpred~\cite{zhang2022d2tpred} treats the signal-induced discontinuity as an explicit dependency in the trajectory model. KI-GAN~\cite{kigan2024} injects intersection-specific knowledge into a generative adversarial trajectory predictor. At unsignalized intersections, \cite{trentin2023multimodal} produces multi-modal interaction-aware predictions, and the macro-micro hierarchical attention method MMH-STA~\cite{sun2023mmhsta} addresses roundabouts. Generative trajectory prediction at intersections was explored earlier through interaction-based GANs~\cite{roy2019gan}. IntersectionNet~\cite{intersectionnet} unrolls trajectories in curvilinear coordinates with geometry-aware attention. All of these are short-horizon predictors evaluated against ground-truth alternative futures; none performs the closed-loop, full-actor-set rollout for thousands of timesteps that we use to evaluate Enactor in the simulator regime.

\textbf{Scene-level world models for autonomous driving.} A third line of work moves beyond per-actor prediction to scene-level generative simulation, including TrafficBots~\cite{trafficbots2023} on the Waymo Open Motion Dataset~\cite{ettinger2021large}, the graph-transformer foundation model of~\cite{wang2023buildingtransportationfoundationmodel} instantiated as TransWorldNG~\cite{transworldng2023}, the city-scale SceneDiffuser++~\cite{tan2025scenediffusercityscaletrafficsimulation}, and the world-model-as-simulator benchmark of~\cite{beyondsimulation2025}. These works model the agents \emph{around} an AV ego vehicle; we instead simulate the entire signalized intersection without a privileged ego, with signal compliance, lane geometry, and intersection-aware safety metrics as first-class concerns.

The work most directly comparable to ours is IntTrajSim~\cite{ranjan2025}, which we use as a baseline. IntTrajSim shares the closed-loop intersection-simulation framing and uses an attention-based architecture, but operates in Cartesian coordinates and applies a post-processing step to enforce stop-line behavior. \cite{metrics} develops an intersection-specific evaluation framework that demonstrates that low trajectory reconstruction error does not imply rule-compliant behavior such as the avoidance of red-light violations or unsafe time-to-collision events; we adopt and extend that metric set in Section~\ref{sec:exp_sim}. \cite{kuncheria2025} has begun exploring richer behavior representations specifically for intersection microsimulation.

In summary, our novelty is not the use of a transformer or of attention but the combination of a geometry-aligned polar state representation, the inclusion of leader-rear-bumper geometric cues as explicit input features, and a closed-loop training and evaluation protocol designed for intersection microsimulation.

\section{Problem Definition}
\label{sec:problem}

We formulate two related tasks. The \emph{simulator task} is the closed-loop generation of next-state distributions for every dynamic actor in an intersection scene, repeated recursively to produce long-horizon trajectories. The \emph{real-world task} is the single-pass prediction of an actor's future positions over a short horizon given a fixed history. The two tasks share an architecture and a training objective; they differ in evaluation, because real-world recordings capture a single realization of the world and cannot be replayed in closed loop against a ground-truth alternative future.

\paragraph*{Notation} An intersection scene contains a set of actors comprising \emph{static actors}, namely lane centerlines, lane boundaries, and other infrastructural elements, and \emph{dynamic actors}, namely vehicles, pedestrians, and the traffic signal phases. In this paper, target dynamic actors are vehicles. Pedestrians, when present in the real-world data, are included as contextual neighbors --- their state contributes to each vehicle's spatial-attention context --- but they are not evaluated as prediction targets and the model is not trained to make claims about pedestrian behavior. We adopt an actor-centric perspective: the future state of each target dynamic actor is predicted from its own history together with the state of its neighbors, the local map, and the local signal. Following~\cite{montali2023} we formulate the problem as a hidden Markov model
\[
\mathcal{H} = \langle \mathcal{S}, \mathcal{O}, p(o_t \mid s_t), p(s_t \mid s_{t-1}) \rangle,
\]
where $\mathcal{S}$ is the set of true world states, $\mathcal{O}$ is the set of observations, $p(o_t \mid s_t)$ is the emission distribution, and $p(s_t \mid s_{t-1})$ governs the latent transition. The marginal observation dynamics are
\[
p_{\mathrm{world}}(o_t \mid s_{t-1}) =
\mathbb{E}_{p(s_t \mid s_{t-1})\, p(o_t \mid s_t)}.
\]
The goal is to learn an approximation $q_{\mathrm{world}}(o_t \mid o^{c}_{<t})$, where the context
\[
o^{c}_{<t} = [\mathrm{omap},\, \mathrm{osignals},\, o_{t-H-1}, \ldots, o_{t-1}]
\]
consists of the static map, signal observations, and the past $H$ observations.

For the actor-centric formulation we model
\begin{equation}
q_{\mathrm{agent}}(o^i_t \mid o^{c(i)}_{<t}),
\qquad
o^{c(i)}_t = [\, \mathrm{map}^i_t,\, \mathrm{signal}^i_t,\, o^i_t,\, n^{1..N(i,t)}_t \,],
\end{equation}
where $o^i_t$ is the state of actor $i$, $n^j_t$ is the state of its $j$-th neighbor, and $N(i,t)$ is the number of neighbors. Following~\cite{zhang2025} we use a polar representation that aligns with the curved trajectories typical at intersections:
\begin{equation}
o^i_t = (r,\, \sin\theta,\, \cos\theta,\, s,\, \sin\alpha,\, \cos\alpha),
\end{equation}
with $r$ the radial distance from the intersection center, $\theta$ the angular position, $s$ the linear speed, and $\alpha$ the heading.

For static elements we use the polyline-based HD-map representation of~\cite{gao2020}: $\mathrm{map}^i_t \in \mathbb{R}^{P^i_t \times L \times 6}$, where $P^i_t$ is the number of polylines near actor $i$ and $L$ is the number of vectors per polyline; each vector is encoded as $(r, \sin\theta, \cos\theta, \ell, \sin\alpha, \cos\alpha)$ with $\ell$ the segment length and $\alpha$ its orientation. These polylines also serve as reference trajectories: by learning the correlation between an actor's state and the nearest reference polyline, the model reduces off-road excursions on turns. The traffic signal is represented as the lane-end vector with the one-hot signal state appended, $(r, \sin\theta, \cos\theta, \ell, \sin\alpha, \cos\alpha, \mathbf{e}_{\mathrm{sig}})$, where $\mathbf{e}_{\mathrm{sig}}\in\{0,1\}^{K}$ encodes the current phase ($K=4$ for red / yellow / flashing-yellow / green).

\begin{figure*}[t]
    \centering
    \includegraphics[width=\textwidth]{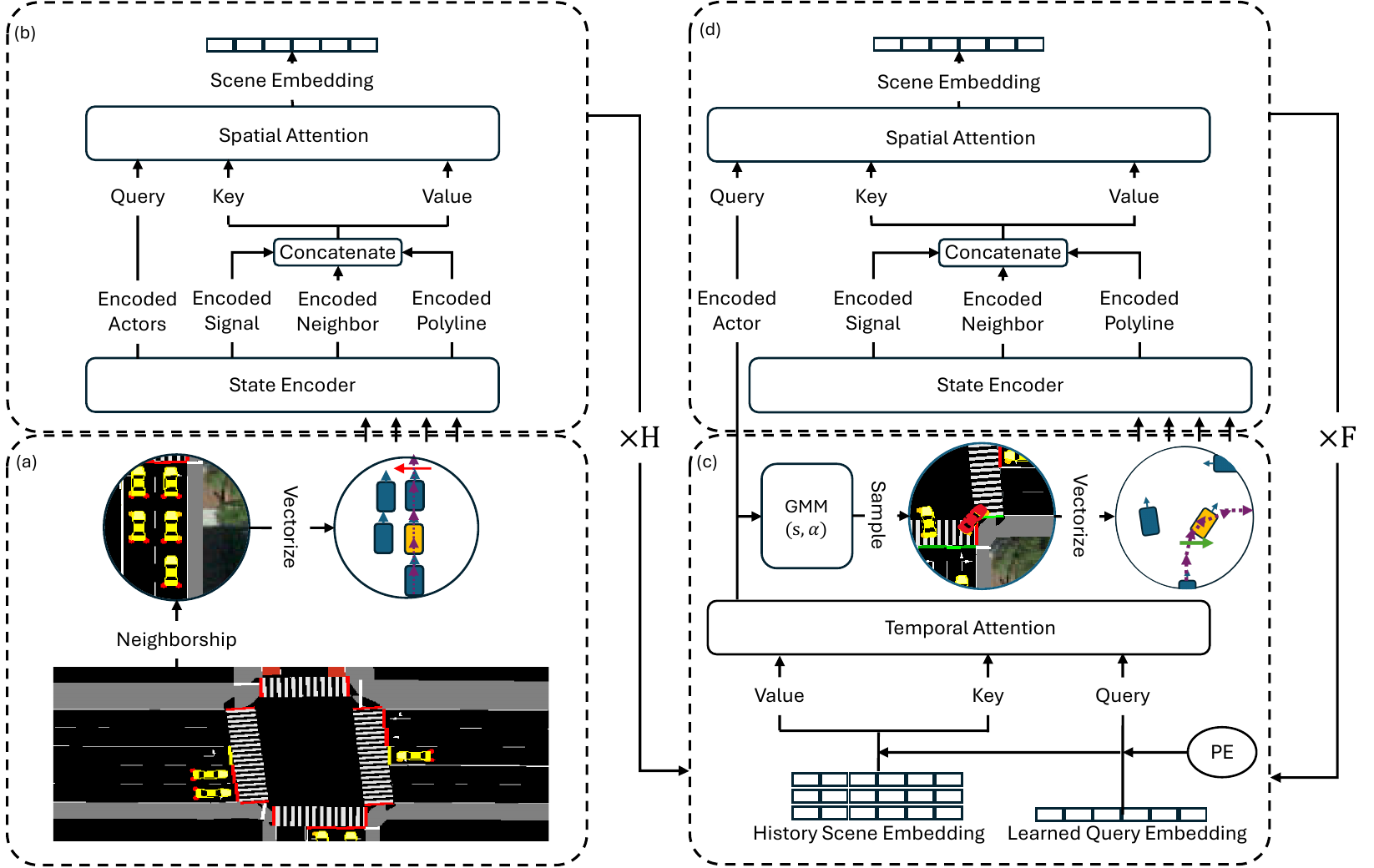}
    \caption{Transformer-based architecture for learning multi-actor interactions and generating trajectories.
    (a)~SUMO initializes the traffic scene from a set of control parameters; for each vehicle, neighboring actors are identified and their states vectorized.
    (b)~A state encoder maps the actor and its neighbors into embeddings, processed by a spatial attention module to produce interaction-aware spatial embeddings over the previous $H$ timesteps.
    (c)~The sequence of spatial embeddings is combined with a learned query embedding and passed through a temporal attention module that parameterizes a distribution over the actor's next-step $(s, \alpha)$. At inference time this distribution is sampled and the next position is obtained by a deterministic kinematic update.
    (d)~The newly sampled state, together with updated neighbor states, is re-encoded and fed back, enabling recursive trajectory unrolling in closed loop.}
    \label{fig:arch}
\end{figure*}

\paragraph*{Prediction target} The model outputs a mixture-of-Gaussians distribution over the actor's next-step speed ($s$) and change in heading ($\alpha$). At inference time, this distribution is sampled and the actor's next position $(r, \theta)$ is computed by a deterministic kinematic update from the sampled $(s, \alpha)$ and the current state (Fig.~\ref{fig:arch}). Concretely, with $(x_t, y_t)$ the Cartesian position derived from $(r_t, \theta_t)$, and $\Delta t$ the timestep,
\begin{equation}
x_{t+1} = x_t + s_{t+1}\,\Delta t\,\cos\alpha_{t+1}, \quad y_{t+1} = y_t + s_{t+1}\,\Delta t\,\sin\alpha_{t+1},
\end{equation}
and the next polar state is $r_{t+1} = \sqrt{x_{t+1}^2 + y_{t+1}^2}$, $\theta_{t+1} = \mathrm{atan2}(y_{t+1}, x_{t+1})$. We predict $\alpha$ directly in radians, without explicit wrapping to $[-\pi, \pi]$; at $10$~fps the frame-to-frame heading change is small enough that wrapping does not arise in practice. The number of mixture components is a hyperparameter (Table~\ref{tab:parameters_2}); empirically, a small mixture suffices for the simulator regime where the data distribution is narrower, while a larger mixture is needed for the real-world regime to capture multimodal behavior. Trajectory-reconstruction metrics (ADE, FDE) are computed in metric Cartesian coordinates after converting each rolled-out polar state to a position in the intersection-centered ground frame.

\paragraph*{Connection to training and evaluation} We train the per-actor model by minimizing the negative log-likelihood of the ground-truth next-step $(s, \alpha)$ under $q_{\mathrm{agent}}(\cdot \mid o^{c(i)}_{<t})$, which is a per-actor surrogate for the marginal $p_{\mathrm{world}}$ above. The closed-loop curriculum of Section~\ref{sec:arch} narrows the gap between conditioning on observed context during training and conditioning on the model's own samples at inference --- the main failure mode of this approximation. The closed-loop simulator-in-the-loop evaluation of Section~\ref{sec:exp_sim} also goes beyond the standard fixed-cohort, fixed-prompt-time setup: it lets the scene graph evolve, with actors entering and leaving the intersection, which is closer to the marginal $p_{\mathrm{world}}$ above, and it covers a much longer time range, so the evaluation reflects how the scene evolves rather than averaging per-actor errors at a single horizon.

\section{Architecture}
\label{sec:arch}
The Enactor architecture has two attention blocks, applied in sequence: a spatial block that captures actor--actor and actor--map interactions at each timestep, and a temporal block that aggregates the spatial embeddings over the history window into a distribution over the next-step $(s, \alpha)$. The spatial-then-temporal decomposition is an explicit choice; it is more tractable than fully joint attention over space and time, and it allows the spatial block to be reused identically during closed-loop unrolling at inference time.

\subsection{Per-Actor Input}
At each timestep $t$, the input for actor $i$ comprises (i)~the actor's own polar state $o^i_t$; (ii)~the states of the $N$ nearest dynamic neighbors within a fixed radius, padded and masked when fewer than $N$ are present; (iii)~the local map polylines within a fixed radius; and (iv)~the lane-end signal vectors for the actor's current and adjacent lanes. The actor's input additionally includes the polar coordinates of the rear bumper of the leading vehicle on the actor's lane, when one exists; the contribution of this feature is isolated by an ablation in Section~\ref{sec:exp_sim}.

\subsection{Spatial Block}
We embed each input element with a multilayer perceptron into a common embedding dimension. The actor's embedding serves as the query in a self-attention layer~\cite{vaswani2017} with the embeddings of its neighbors, map polylines, and signal vectors as keys and values. The output is an interaction-aware embedding for the actor at each timestep in the history.

\subsection{Temporal Block}
The sequence of $H$ spatial embeddings is concatenated with a learned query embedding and passed through a temporal attention module that produces a single embedding summarizing the actor's history. This embedding parameterizes the distribution over the actor's next-step $(s, \alpha)$.

\subsection{Closed-Loop Training}
The training loss is the negative log-likelihood of the ground-truth next-step $(s, \alpha)$ under the predicted distribution. To mitigate covariate shift during closed-loop rollout, we use a curriculum schedule: the model receives ground-truth context for the first portion of training, and ground-truth context is progressively replaced by the model's own samples for an increasing fraction of timesteps. We use the AdamW optimizer with a cosine learning-rate schedule and gradient clipping.

\paragraph*{Hyperparameters} A comprehensive list of hyperparameters used during training is provided in Table~\ref{tab:parameters_1}, with architecture sizes in Table~\ref{tab:parameters_2}. We use batch size $256$, $4000$ warmup steps for the learning-rate scheduler, and dropout ratio $0.1$. The same architecture is used for both the simulator and real-world regimes; the regimes differ only in the number of mixture components used in the output head ($5$ for the simulator regime, $25$ for the real-world regime; see Section~\ref{sec:problem}).
\begin{table*}[t] 
    \centering
    \scriptsize
    \setlength{\tabcolsep}{3pt}
    \renewcommand{\arraystretch}{1.2}
    \begin{tabularx}{\textwidth}{l *{10}{>{\centering\arraybackslash}X}} 
        \rowcolor{black!30}
        \textbf{Model} &
        \textbf{History Length} &
        \textbf{Prediction Horizon} &
        \textbf{Radius Normalization Factor} &
        \textbf{Speed Normalization Factor} &
        \textbf{Max Neighbors} &
        \textbf{Neighbor Radius} &
        \textbf{Max Polyline} &
        \textbf{Polyline Radius} &
        \textbf{Signal Radius} \\
        \rowcolor{gray!15}
        Enactor   & 20 & 20 & 50  & 10  & 10 & 50 & 10 & 20 & 20 \\
    \end{tabularx}
    \caption{Data-processing parameters for training on simulated and real-world data.}
    \label{tab:parameters_1}
\end{table*} 

\begin{table*}[t] 
    \centering
    \scriptsize
    \setlength{\tabcolsep}{3pt}
    \renewcommand{\arraystretch}{1.2}
    \begin{tabularx}{\columnwidth}{l *{6}{>{\centering\arraybackslash}X}} 
        \rowcolor{black!30}
        \textbf{Model} &
        \textbf{Embedding Size} &
        \textbf{Spatial Attention Heads} &
        \textbf{Spatial Attention Layers} &
        \textbf{Temporal Attention Heads} &
        \textbf{Temporal Attention Layers} &
        \textbf{GMM Modes} \\
        \rowcolor{gray!05}
        Enactor  & 256 & 4 & 8 & 4 & 8 & 5 / 25\\
    \end{tabularx}
    \caption{Enactor architecture parameters for training on simulated and real-world data. The same architecture is used in both regimes; only the number of mixture components in the GMM head over $(s, \alpha)$ differs --- $5$ for the simulator regime, $25$ for the real-world regime (shown as ``$5$~/~$25$'' in the table).}
    \label{tab:parameters_2}
\end{table*} 

\begin{table}[t]
    \centering
    \scriptsize
    \setlength{\tabcolsep}{3pt}
    \renewcommand{\arraystretch}{1.2}
    \begin{tabularx}{\columnwidth}{l *{7}{>{\centering\arraybackslash}X}}
        \rowcolor{black!30}
        \textbf{Data Source} &
        \textbf{East Leg(/H)} &
        \textbf{South Leg(/H)} &
        \textbf{West Leg(/H)} &
        \textbf{North Leg(/H)} &
        \textbf{Unmapped (/H)} &
        \textbf{Test Train Split} \\
        \rowcolor{gray!15}
        Simulated   & 307 & 323 & 289 & - & - & 0.9\\
        \rowcolor{gray!05}
        Real World  & 570 & 831 & 522 & 720 & 1504 & 0.9\\
    \end{tabularx}
    \caption{Traffic Data Description}
    \label{tab:data_description_1}
\end{table}

\begin{table}[t]
    \centering
    \scriptsize
    \setlength{\tabcolsep}{3pt}
    \renewcommand{\arraystretch}{1.2}
    \begin{tabularx}{\columnwidth}{l *{7}{>{\centering\arraybackslash}X}}
        \rowcolor{black!30}
        \textbf{Data Source} &
        \textbf{Train (Number of Actors)} &
        \textbf{Test (Number of Actors)} &
        \textbf{Train (Seconds)} &
        \textbf{Test (Seconds)} \\
        \rowcolor{gray!15}
        Simulated   & 4959 & 601 & 19440 & 2160 \\
        \rowcolor{gray!05}
        Real World  & 3386 & 922 & 3240 & 360 \\
    \end{tabularx}
    \caption{Train/Test Split Description}
    \label{tab:data_description_2}
\end{table}

\section{Data Sources}
\label{sec:data}
We use two complementary data sources: SUMO simulation data for closed-loop training and evaluation, and a fish-eye-camera deployment for the real-world study.

\subsection{Simulator Data}
\label{sec:data_sim}
\begin{figure*}[t]
    \centering
    \includegraphics[width=\textwidth]{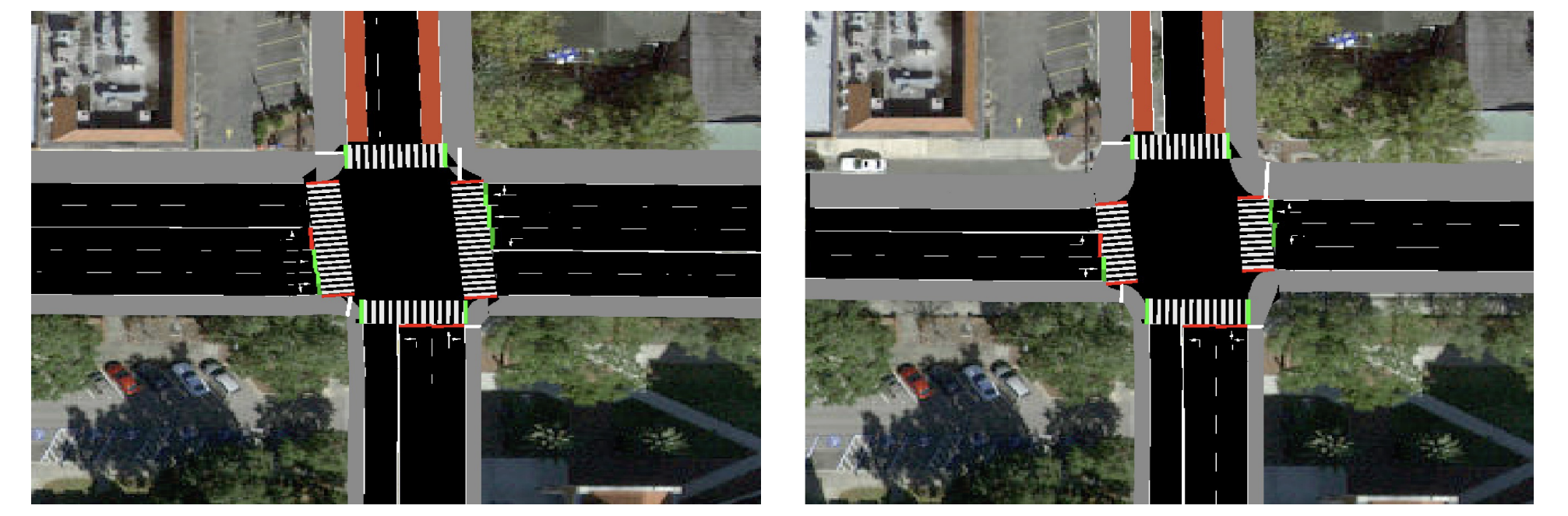}
    \caption{Digital representations of the two simulated intersections. left: SUMO representation of the intersection at West University. right: The same intersection with the middle through-lanes of the east- and westbound approaches removed.}
    \label{fig:intersections}
\end{figure*}

We use SUMO to generate trajectory data for two intersection geometries. The first is a digital reconstruction of the intersection at West University Avenue and NW 17th Street in Gainesville, Florida; the second is the same intersection with the middle through-lanes of the eastbound and westbound approaches removed. The two geometries are shown in Fig.~\ref{fig:intersections}. SUMO is configured with calibrated arrival rates per approach and a fixed-time signal plan; the simulator generates vehicular traffic only, consistent with the vehicle-centric scope established in Section~\ref{sec:intro}. We simulate $6$~hours ($21{,}600$~s) of traffic at a temporal resolution of $0.1$~s, yielding approximately $1{,}241{,}600$ training exemplars. Table ~\ref{tab:data_description_1} describes the number of vehicle arriving per hour at each leg of the intersection, for both simulated and real world dataset. As seen in ~\ref{fig:intersections}, the intersection at West University Avenue does not allow incoming traffic from the north leg which we have recreated in the digital map inside SUMO. The real world dataset has trajectories which are incomplete due to false detections and missed tracks. The trajectories either abruptly start or end in the middle of the intersection. We do not map these vehicles to a cluster and use these vehicles only as neighbors of other vehicles with valid trajectories.

\subsection{Real-World Data and Preprocessing}
\label{sec:data_real}
The real-world data are collected from a fish-eye camera installed at a signalized intersection adjacent to the University of Florida campus. The video stream is processed by a YOLO-based detection-and-tracking pipeline developed within our laboratory, which produces per-frame bounding boxes and persistent identities for vehicles and pedestrians. We use one hour of video for the experiments reported in Section~\ref{sec:exp_real}. Lane-wise signal phases are recovered from the same video by visual inspection of the traffic movement.

The raw trajectories returned by the pipeline contain two characteristic noise sources. The first is intermittent missing frames for individual actors, caused by short-lived detection failures. The second, and more consequential, is lateral drift along the direction perpendicular to the actor's motion, caused by jitter in the bounding-box centroid that the tracking step cannot fully suppress. Without correction, the resulting position sequences exhibit artificially large velocities and accelerations whenever the centroid steps sideways between consecutive frames; this is most pronounced on curved paths, where the lateral noise is locally tangent to the trajectory's true curvature and is therefore difficult to distinguish from real heading change. Training on uncorrected trajectories therefore exposes the model to higher-frequency artifacts in the dynamics signal; whether this hurts the learned predictor enough to warrant correction is the empirical question we test in Section~\ref{sec:exp_real}.

\begin{figure}[t]
    \centering
    \includegraphics[width=\columnwidth]{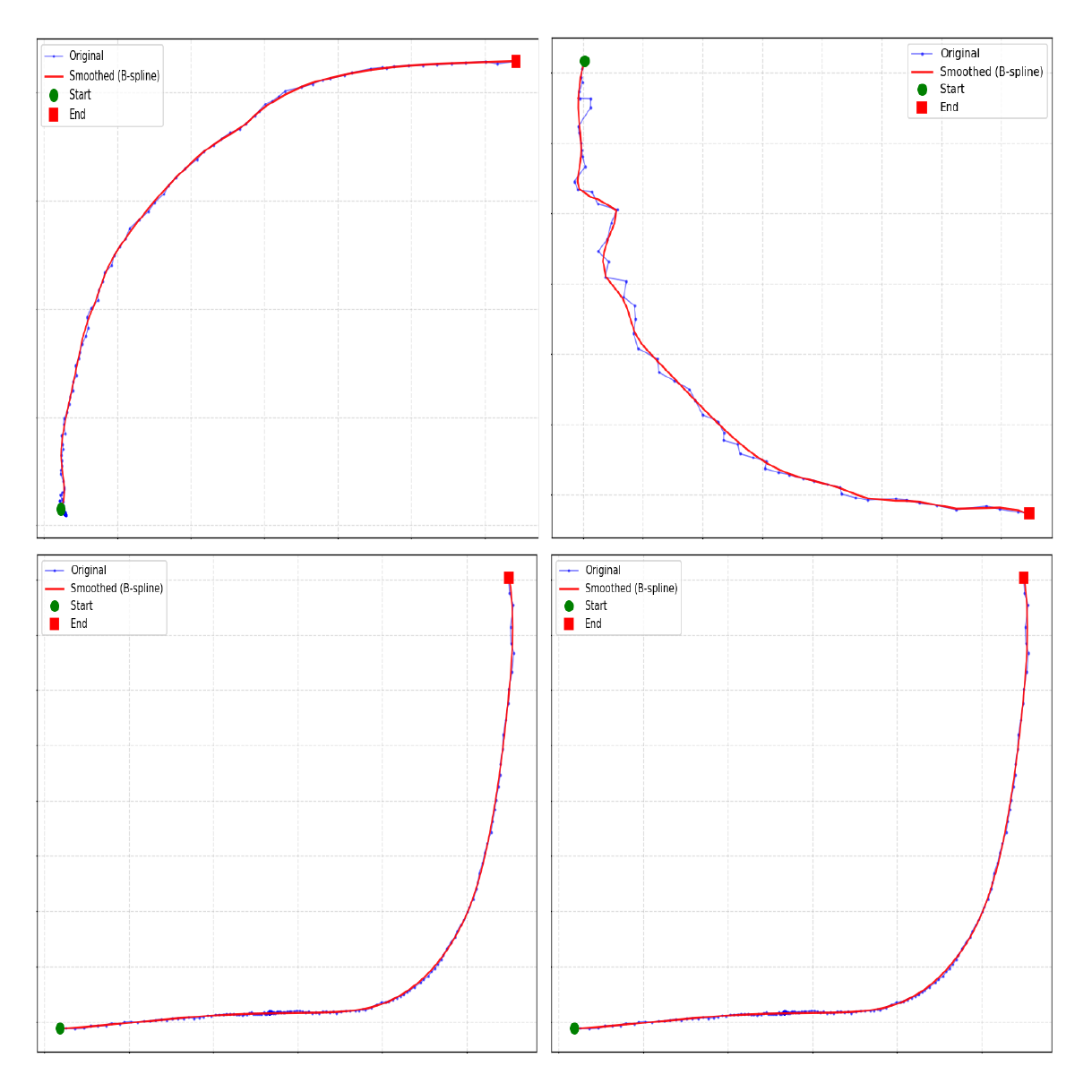}
    \caption{Lateral drift from detection error is corrected by fitting a smoothing spline to each actor's trajectory and projecting the observed positions onto the spline.}
    \label{fig:smooth}
\end{figure}

We fill missing frames by linear interpolation between the last and next observed positions of the actor. We address lateral drift by fitting a smoothing spline to each actor's trajectory in a metric ground frame and projecting each observed position onto the spline. Fig.~\ref{fig:smooth} shows the effect: the raw trace has visible lateral oscillation around the lane centerline, which the spline projection removes while leaving the longitudinal motion alone. This step is about measurement error, not modeling. The YOLO bounding-box pipeline adds lateral jitter that is not part of the actor's true motion, and we test whether removing it gives the downstream model a cleaner dynamics signal. Whether smoothing actually helps a learned trajectory predictor is itself an empirical question. A smoothing spline acts as a low-pass filter, so it should help short-horizon prediction --- where centroid jitter is uninformative noise the model should not have to fit --- and may hurt longer-horizon prediction, where the small bias smoothing introduces in the current state can compound. We report both Enactor-Smooth (the smoothed variant) and Enactor-Raw (the unsmoothed variant) in Section~\ref{sec:exp_real} and discuss the trade-off there.

Enactor-Smooth should be interpreted as a training-input denoising variant rather than an online preprocessing pipeline. Enactor-Raw and Enactor-Smooth are evaluated on the same held-out raw vehicle histories and raw future targets, so the comparison isolates the effect of smoothing the training input. To avoid information leakage across the chronological train/test split, spline fitting for Enactor-Smooth is performed only on trajectories within the training segment; held-out future positions are not used in constructing smoothed training inputs. A causal smoother would be needed to support an online-deployment claim; we do not make one.

\begin{table}[t]
    \centering
    \scriptsize
    \setlength{\tabcolsep}{3pt}
    \renewcommand{\arraystretch}{1.2}
    \begin{tabularx}{\columnwidth}{l *{4}{>{\centering\arraybackslash}X}}
        \rowcolor{black!30}
        \textbf{Data} &
        \textbf{Median displacement (m)} &
        \textbf{95th-percentile displacement (m)} &
        \textbf{Median delta head (radian)} &
        \textbf{95th-percentile delta head (radian)}\\
        \rowcolor{gray!15}
        Raw  & 0.0   & 1.2292 & 0.0175 & 2.7255 \\
        \rowcolor{gray!05}
        Smoothed       & 0.0 & 1.2288  & 0.0006 & 3.0845\\
        \rowcolor{gray!15}
        Change         & 0.0   & 0.0004 & 0.0169 & -0.359 \\
    \end{tabularx}
    \caption{Quantifying the effect of spline smoothing on trajectory measurements. The median per-frame displacement is reported as $0.0$ because a substantial fraction of frames belong to stationary vehicles (e.g., halted at a red phase), which dominate the median on both raw and smoothed traces. The dominant effect of smoothing is on the median frame-to-frame heading change, which drops by roughly $97\%$ ($0.0175 \to 0.0006$~rad); displacement statistics are essentially unchanged, and the $95$th-percentile heading change is not improved by smoothing.}
    \label{tab:smoothing_1}
\end{table}

\begin{figure*}[t]
    \centering
    \includegraphics[width=\textwidth]{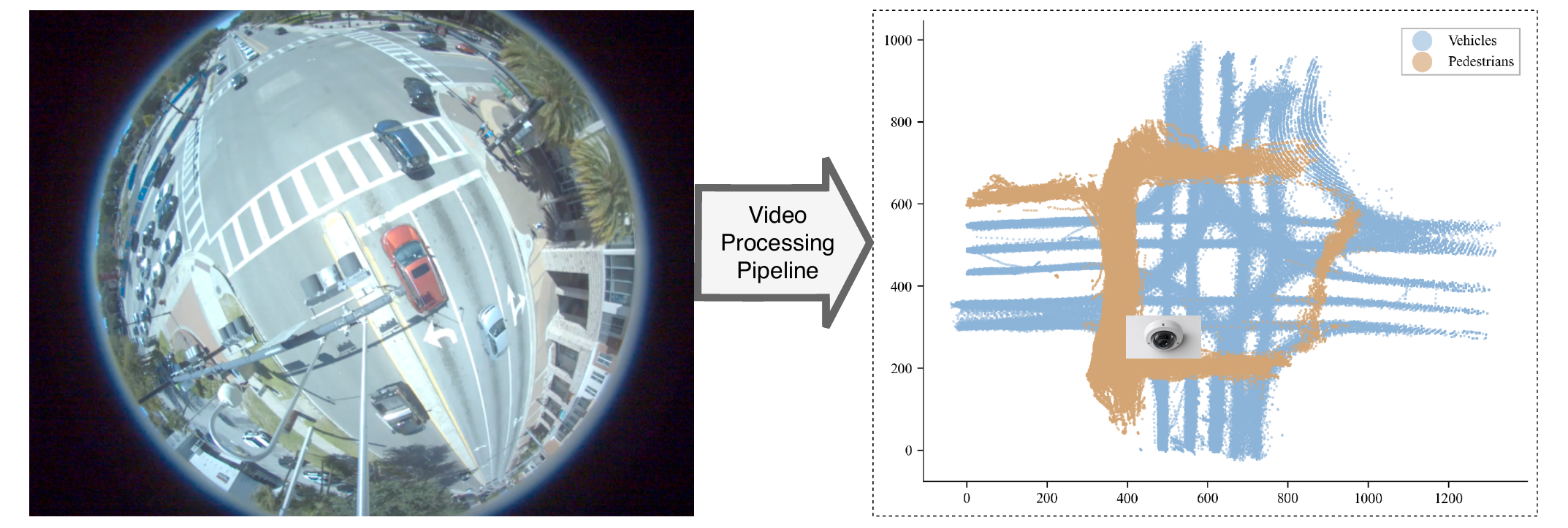}
    \caption{Traffic video is collected through fish-eye camera at a frame rate of 10 frames per second, and processed though a data processing pipeline that detects and tracks vehicles and pedestrians. The processed trajectories are mapped to the the Google Maps coordinates to generate final trajectories.}
    \label{fig:processing}
\end{figure*}

Trajectories are smoothed using a parametric B-spline (\texttt{scipy.splprep}, cubic). Control points are selected greedily at a minimum spacing of $1$~m; the smoothing parameter is set to $s = n$, where $n$ is the number of control points, following the standard SciPy heuristic of one residual unit per knot. Each original observation is then projected to its nearest point on the densely sampled spline.
Data is collected at a frame rate of $10$~frames per second at a resolution of $1280 \times 960$ pixels~\cite{jimaging8040101}. We use one hour of recorded data: the first $54$ minutes are used for training and the final $6$ minutes are held out for testing. Tables~\ref{tab:data_description_1} and~\ref{tab:data_description_2} report per-leg vehicle arrival rates and the corresponding training and test split sizes. The data-processing pipeline uses thin-plate splines to map the fish-eye video into a rectangular ground frame, by matching the fish-eye image coordinates of lane centerlines to their corresponding Google Maps coordinates. Both vehicles and pedestrians are detected and tracked by the pipeline; pedestrian trajectories enter the model as neighbor context for nearby vehicles, but are not used as evaluation targets, consistent with the scope established in Section~\ref{sec:intro}.


\section{Experiments and Results: Simulator Regime}
\label{sec:exp_sim}

We evaluate Enactor on the SUMO data of Section~\ref{sec:data_sim} in a closed-loop ``simulation-in-the-loop'' framework. After establishing the simulator results we report the corresponding real-world study in Section~\ref{sec:exp_real}.

\paragraph*{Purpose of the simulator regime} The simulator regime is a controlled testbed for closed-loop learned actor dynamics, not a claim that Enactor should replace SUMO. We use it to test whether a learned model can stay stable and rule-aware when it has to control all dynamic vehicles over a continuously refreshing actor set against a ground truth that can be replayed --- something the real-world data cannot offer, since each recording is a single realization of the world. The real-world study in Section~\ref{sec:exp_real} then tests whether the same architecture can be trained and evaluated on naturalistic trajectories.

\subsection{Closed-Loop Evaluation Protocol}
SUMO generates vehicular traffic at a calibrated arrival rate and controls the signal phases. For each vehicle, the first few timesteps are governed by SUMO's built-in car-following model. After this warm-start period, Enactor takes over and unrolls trajectories in closed loop for $40{,}000$ steps ($4{,}000$~s).

During the rollout we simulate trajectories of $1{,}026$ distinct actors, with roughly $7$ concurrent at any given time and a mean residence time of about $28$~seconds.

\paragraph*{A note on simulation speed} Simulation speed is not a focus of this paper, and we report no runtime numbers here; our claims concern actor-actor interaction realism, not rollout throughput.

\subsection{Metrics}
\label{sec:metrics_sim}
We use three groups of metrics: trajectory reconstruction, aggregate flow, and intersection-aware safety.

\subsubsection*{Trajectory reconstruction}
Trajectory reconstruction metrics compare the predicted trajectories against ground truth. We use \textbf{ADE / FDE / NLL}: average and final displacement error in Cartesian coordinates, and the negative log-likelihood of the ground-truth $(s, \alpha)$ sequence under the model's predicted distribution. Results are in Tables~\ref{tab:safety_sim_-1} and~\ref{tab:safety_sim_0} for the two intersections.

\begin{table}[t]
    \centering
    \scriptsize
    \setlength{\tabcolsep}{3pt}
    \renewcommand{\arraystretch}{1.2}
    \begin{tabularx}{\columnwidth}{l *{5}{>{\centering\arraybackslash}X}}
        \rowcolor{black!30}
        \textbf{Model} &
        \textbf{NLL} &
        \textbf{ADE @ 2s} &
        \textbf{FDE @ 2s} \\
        \rowcolor{gray!15}
        IntTrajSim         & -3.7   & 0.183 & 0.462  \\
        \rowcolor{gray!05}
        Enactor       & -7.12 & 0.137  & 0.279 \\
    \end{tabularx}
    \caption{Trajectory Reconstruction metrics at the first intersection. NLL is computed under each model's own predicted distribution and is not directly comparable across models; ADE/FDE are the primary cross-model comparison.}
    \label{tab:safety_sim_-1}
\end{table}

\begin{table}[t]
    \centering
    \scriptsize
    \setlength{\tabcolsep}{3pt}
    \renewcommand{\arraystretch}{1.2}
    \begin{tabularx}{\columnwidth}{l *{5}{>{\centering\arraybackslash}X}}
        \rowcolor{black!30}
        \textbf{Model} &
        \textbf{NLL} &
        \textbf{ADE @ 2s} &
        \textbf{FDE @ 2s} \\
        \rowcolor{gray!15}
        IntTrajSim         & -4.30   & 0.122 & 0.309  \\
        \rowcolor{gray!05}
        Enactor       & -7.17 & 0.128  & 0.269 \\
    \end{tabularx}
    \caption{Trajectory Reconstruction metrics at the second intersection. NLL is computed under each model's own predicted distribution and is not directly comparable across models; ADE/FDE are the primary cross-model comparison.}
    \label{tab:safety_sim_0}
\end{table}

\subsubsection*{Intersection-aware safety}
\begin{itemize}
    \item \textbf{Red-light violations:} crossings of the stop bar during a red phase.
    \item \textbf{Mid-intersection low-speed event:} speed of $8$~kmph or less for $2$~seconds inside the intersection box, outside of the warm-start period.
    \item \textbf{Pre-stopbar stoppage:} stops more than a vehicle length before the stop bar on a green phase.
    \item \textbf{Unsafe Deceleration:} timesteps with longitudinal Deceleration outside a physically plausible range of -0.47g ($g = 9.8\,\mathrm{m/s^2}$)
    \item \textbf{TTC events:} time-to-collision events below a safety threshold of 1 second.
\end{itemize}

\subsubsection*{Aggregate flow}
\begin{itemize}
    \item \textbf{Average speed:} mean and standard deviation of a Gaussian fit to the distribution of instantaneous vehicle speeds.
    \item \textbf{Average travel time:} mean and standard deviation of travel times for vehicles traversing the intersection.
\end{itemize}

\subsection{Results: First Intersection}
We compare Enactor against the IntTrajSim baseline~\cite{ranjan2025}. The story below has three parts that should be read together: aggregate-flow realism, where the model is strong; rule compliance, where it improves on the baseline once the leader-rear-bumper feature is included; and close-range interaction realism, which remains an open problem.

\paragraph*{Aggregate flow}
On both speed and travel-time distributions, Enactor matches the ground-truth distributions substantially more closely than the baseline (Tables~\ref{tab:speed_sim_1} and~\ref{tab:tt_sim_1}). On travel time, KL divergence is reduced by over an order of magnitude; on speed, it is reduced by roughly $5\times$. The macroscopic distributions of vehicles produced by closed-loop rollout for $4{,}000$~s are therefore close to those produced by SUMO itself.

\paragraph*{Rule compliance}
Table~\ref{tab:safety_sim_1} reports the intersection-aware safety metrics. We include an ablation, denoted ``Enactor (no leader)'', that strips the leader rear-bumper feature from the actor's input but is otherwise identical to Enactor. Without the feature, the model is worse than the baseline on stop-line and TTC metrics: $199$ red-light violations and $577$ TTC events versus $169$ and $378$ for the baseline. With the feature included, red-light violations fall to $5$ and TTC events fall to $366$, while the unsafe-acceleration count is unchanged. This is a meaningful gain on rule compliance specifically.

The behavior of the ablation is consistent with a qualitative observation made during inspection: without an explicit geometric reference for the rear of the leading vehicle, the model decelerates appropriately when approaching a red signal or a leading vehicle but does not reliably stop at the correct location, instead overshooting the stop line or the rear of the leader. The leader rear-bumper feature provides exactly that geometric reference and is the single change with the largest effect on Table~\ref{tab:safety_sim_1}.

\paragraph*{A note on baseline post-processing}
IntTrajSim is reported in two configurations: a vanilla configuration that uses only the model's raw predictions and a post-processed configuration that adds an explicit stop-line enforcement step. Enactor is reported without analogous post-processing. With its post-processing in place, IntTrajSim reaches $169$ red-light violations and $378$ TTC events at Site~1; with the post-processing removed, the same model reaches $175$ red-light violations and $258$ TTC events. The post-processing therefore makes only a small difference to the safety metrics for IntTrajSim. Under the same protocol, Enactor reduces red-light violations by more than an order of magnitude relative to either IntTrajSim configuration ($5$ vs.\ $169$/$175$), and we read the gap as a measure of relative safety realism within the closed-loop regime that both models occupy.

\paragraph*{Close-range interaction realism remains limited}
Even with the leader-rear-bumper feature, the TTC count remains higher than the count obtained from raw SUMO data, indicating that close-range interactions are not yet modeled at full fidelity. We name this as the principal remaining limitation of the model and return to it in Section~\ref{sec:conclusion}.

\begin{table}[t]
    \centering
    \scriptsize
    \setlength{\tabcolsep}{3pt}
    \renewcommand{\arraystretch}{1.2}
    \begin{tabularx}{\columnwidth}{l *{5}{>{\centering\arraybackslash}X}}
        \rowcolor{black!30}
        \textbf{Model} &
        \textbf{Red Light Viol.} &
        \textbf{Mid-Int.\ Low-Speed} &
        \textbf{Pre-Stopbar Stop.} &
        \textbf{Unsafe Accel.} &
        \textbf{TTC Event} \\
        \rowcolor{gray!15}
        Raw SUMO         & 0   & 44 & 0  & 0  & 32  \\
        \rowcolor{gray!05}
        IntTrajSim (vanilla)      & 175 & 0  & 3  & 0 & 258 \\
        \rowcolor{gray!05}
        IntTrajSim (post-process)      & 169 & 1  & 0  & 11 & 378 \\
        \rowcolor{gray!15}
        Enactor (no leader) & 199 & 7  & 37 & 6  & 577 \\
        \rowcolor{gray!15}
        Enactor          & \textbf{5} & 16 & 4 & \textbf{6} & \textbf{366} \\
    \end{tabularx}
    \caption{Intersection-aware safety metrics at the first intersection. Bold entries mark the best non-baseline result; raw SUMO is shown as a reference for the metric values produced by the rule-compliant SUMO behavior models themselves.}
    \label{tab:safety_sim_1}
\end{table}

\begin{table}[t]
\centering
\scriptsize
\setlength{\tabcolsep}{3pt}
\renewcommand{\arraystretch}{1.15}
\begin{tabularx}{\columnwidth}{l *{3}{>{\centering\arraybackslash}X}}
    \rowcolor{black!30}
    \textbf{Model} &
    \textbf{Mean (m/s)} &
    \textbf{Std.\ Dev.\ (m/s)} &
    \textbf{KL Divergence} \\
    \rowcolor{gray!15}
    Ground Truth & 2.82 & 3.77 & -     \\
    \rowcolor{gray!05}
    IntTrajSim (vanilla) & 4.81 & 3.89 & 0.1318 \\
    \rowcolor{gray!05}
    IntTrajSim (post-process)   & 3.28 & 3.54 & 0.0112 \\
    \rowcolor{gray!15}
    Enactor (no leader) & 2.87 & 3.59 & \textbf{0.0024} \\
    \rowcolor{gray!15}
    Enactor      & 2.68 & 3.49 & 0.0063 \\
\end{tabularx}
\caption{Speed-distribution statistics at the first intersection.}
\label{tab:speed_sim_1}
\end{table}

\begin{table}[t]
\centering
\scriptsize
\setlength{\tabcolsep}{3pt}
\renewcommand{\arraystretch}{1.15}
\begin{tabularx}{\columnwidth}{l *{3}{>{\centering\arraybackslash}X}}
    \rowcolor{black!30}
    \textbf{Model} &
    \textbf{Mean (s)} &
    \textbf{Std.\ Dev.\ (s)} &
    \textbf{KL Divergence} \\
    \rowcolor{gray!15}
    Ground Truth & 28.19 & 16.26 & -     \\
    \rowcolor{gray!05}
    IntTrajSim (vanilla) & 16.31 & 13.93 & 0.28854 \\
    \rowcolor{gray!05}
    IntTrajSim (post-process)   & 23.92 & 17.53 & 0.04043 \\
    \rowcolor{gray!15}
    Enactor (no leader) & 27.24 & 16.03 & 0.00190 \\
    \rowcolor{gray!15}
    Enactor      & 28.78 & 16.53 & \textbf{0.00093} \\
\end{tabularx}
\caption{Travel-time statistics at the first intersection.}
\label{tab:tt_sim_1}
\end{table}

\subsection{Results: Second Intersection}
We evaluate Enactor on a second intersection geometry, in which the middle through-lanes for the eastbound and westbound approaches are removed. The model is trained from scratch on data collected from this geometry and evaluated under the same closed-loop protocol used for the first intersection.

The aggregate-flow picture mirrors Site 1: Enactor matches the travel-time distribution much more closely than IntTrajSim (Table~\ref{tab:tt_sim_2}), and its speed mean and standard deviation are closer to ground truth (Table~\ref{tab:speed_sim_2}). Safety results are mixed in absolute terms but clear relative to the baseline: Enactor has $1$ red-light violation, $82$ mid-intersection low-speed events, and $534$ TTC events, against IntTrajSim's $278$, $234$, and $794$ under the same protocol (Table~\ref{tab:safety_sim_2}). Both IntTrajSim and Enactor are trained from scratch on Site 2 data using the same protocol as at Site 1. Both sustain the $4{,}000$-second rollout against the dynamic-actor cohort at the modified geometry, so the comparison again measures relative safety realism rather than gross stability. Enactor's $534$ TTC events are still well above the raw-SUMO reference of $42$, so close-range interaction realism remains the main limitation --- just less severe than the baseline's. One regression is specific to the modified geometry: mid-intersection low-speed events rise from $16$ at Site 1 to $82$ at Site 2. Removing the middle through-lanes on the east--west approaches compresses east- and westbound flow into a single through-lane each, and our inspection suggests that Enactor keeps red-light compliance but accelerates more slowly than the ground truth when the signal turns from red to green. That slow movement inside the intersection box is what the metric captures. The phenomenon is more visible at Site 2 because of the queueing dynamics that arise under this lane compression. The failure mode is distinct from red-light violation and points to lane-occupancy or queue-state features as a candidate input for future work.

\begin{table}[t]
    \centering
    \scriptsize
    \setlength{\tabcolsep}{3pt}
    \renewcommand{\arraystretch}{1.2}
    \begin{tabularx}{\columnwidth}{l *{5}{>{\centering\arraybackslash}X}}
        \rowcolor{black!30}
        \textbf{Model} &
        \textbf{Red Light Viol.} &
        \textbf{Mid-Int.\ Low-Speed} &
        \textbf{Pre-Stopbar Stop.} &
        \textbf{Unsafe Accel.} &
        \textbf{TTC Event} \\
        \rowcolor{gray!15}
        Raw SUMO  & 0 & 41 & 3  & 0 & 42  \\
        \rowcolor{gray!15}
        IntTrajSim  & 278 & 234 & 0  & 0 & 794  \\
        \rowcolor{gray!05}
        Enactor   & 1 & 82 & 20 & 4 & 534 \\
    \end{tabularx}
    \caption{Intersection-aware safety metrics at the second intersection.}
    \label{tab:safety_sim_2}
\end{table}

\begin{table}[t]
\centering
\scriptsize
\setlength{\tabcolsep}{3pt}
\renewcommand{\arraystretch}{1.15}
\begin{tabularx}{\columnwidth}{l *{3}{>{\centering\arraybackslash}X}}
    \rowcolor{black!30}
    \textbf{Model} &
    \textbf{Mean (s)} &
    \textbf{Std.\ Dev.\ (s)} &
    \textbf{KL Divergence} \\
    \rowcolor{gray!15}
    Ground Truth & 32.31 & 20.60 & -      \\
    \rowcolor{gray!15}
    IntTrajSim & 22.35 & 21.72 & 0.11895      \\
    \rowcolor{gray!05}
    Enactor      & 33.38 & 20.01 & 0.00217 \\
\end{tabularx}
\caption{Travel-time statistics at the second intersection.}
\label{tab:tt_sim_2}
\end{table}

\begin{table}[t]
\centering
\scriptsize
\setlength{\tabcolsep}{3pt}
\renewcommand{\arraystretch}{1.15}
\begin{tabularx}{\columnwidth}{l *{3}{>{\centering\arraybackslash}X}}
    \rowcolor{black!30}
    \textbf{Model} &
    \textbf{Mean (m/s)} &
    \textbf{Std.\ Dev.\ (m/s)} &
    \textbf{KL Divergence} \\
    \rowcolor{gray!15}
    Ground Truth & 2.52 & 3.69 & -- \\
    \rowcolor{gray!15}
    IntTrajSim & 3.52 & 3.85 & 0.0385 \\
    \rowcolor{gray!15}
    Enactor      & 2.35 & 3.29 & 0.0132 \\
\end{tabularx}
\caption{Speed-distribution statistics at the second intersection.}
\label{tab:speed_sim_2}
\end{table}

\section{Experiments and Results: Real-World Regime}
\label{sec:exp_real}

The real-world setting differs from the simulator in two ways that shape the experimental design. First, each segment of the recording captures a single realization of the world. There is no equivalent to running multiple SUMO seeds, and no ground-truth alternative future to compare a closed-loop rollout against. Second, the input trajectories are noisy estimates of the true motion, run through the preprocessing of Section~\ref{sec:data_real}; the evaluation therefore folds in the residual error of that preprocessing. Both push us away from long closed-loop rollouts and toward shorter, single-pass prediction.

\subsection{Real-World Evaluation Protocol}
We split the recording chronologically into a $90$/$10$ training/test split by total time; the per-segment actor counts and durations are reported in Table~\ref{tab:data_description_2}. On the evaluation segment, the model receives a context of $H$ raw observed timesteps and predicts the next $F$ timesteps for every actor present at the prompt time. Both Enactor-Smooth and Enactor-Raw use the same raw history at test time and predict against the same raw target trajectory; the two variants differ only in whether the training trajectories were spline-smoothed. Predictions are made in a single forward pass per actor; we do not attempt a multi-actor closed-loop rollout, because there is no ground-truth scene against which to compare it. The procedure is repeated across multiple prompt times spanning the held-out segment.

For the evaluation step above, we first identify vehicles whose trajectories are present for the past $H$ steps and the next $F$ timesteps. We score only mapped trajectories; unmapped vehicles enter only as neighbors during spatial-embedding computation, consistent with the vehicle-centric scope of this paper. The number of valid evaluation samples is $60{,}472$ at the $2$-second prediction horizon, $48{,}597$ at $5$ seconds, and $41{,}802$ at $10$ seconds. Because prompts are generated at overlapping timesteps within each actor's track, these counts should be interpreted as prediction instances rather than as statistically independent trajectories.

\subsection{Metrics}
\label{sec:metrics_real}
The single-pass setting calls for a different metric set, focused on the predictive distribution at a few horizons:
\begin{itemize}
    \item \textbf{Multi-horizon FDE:} $\ell_2$ distance between predicted and ground-truth positions at $t+2$~s, $t+5$~s, and $t+10$~s after the prompt.
    \item \textbf{Multi-horizon ADE:} $\ell_2$ distance averaged over the prediction window up to each horizon.
    \item \textbf{Within-threshold success rates:} the fraction of predictions within distance thresholds appropriate to each horizon.
    \item \textbf{Vehicle-only evaluation:} all real-world metrics are reported for mapped vehicle trajectories. Pedestrians are used only as contextual neighbors.
\end{itemize}

\paragraph*{Constant-velocity baselines} We compare against three versions of a constant-velocity baseline, each extrapolating linearly in the metric ground frame from the prompt-time position but estimating velocity over a different window of past history. \emph{CV (last~$1$)} uses the two-frame finite difference between the last two observed positions: $\mathbf{v}_t = (\mathbf{p}_t - \mathbf{p}_{t-1}) / \Delta_{\text{frame}}$. \emph{CV (avg~$10$)} averages the per-frame finite-difference velocities over the final $10$ observed frames ($1.0$~s at $10$~fps). \emph{CV (ls~$10$)} fits a least-squares line to the final $10$ observed positions and uses its slope. The absolute measure of the velocity is kept constant and the head angle is derived from nearest lane center lines that the vehicle traverses. Reporting all three windows shows that Enactor's advantage is not an artifact of picking the noisiest baseline: the more aggressive ls~$10$ variant closes much of the gap to the naive last~$1$ at longer horizons, but Enactor still outperforms it at every horizon evaluated.

\begin{figure}[t]
    \centering
    \includegraphics[width=\columnwidth]{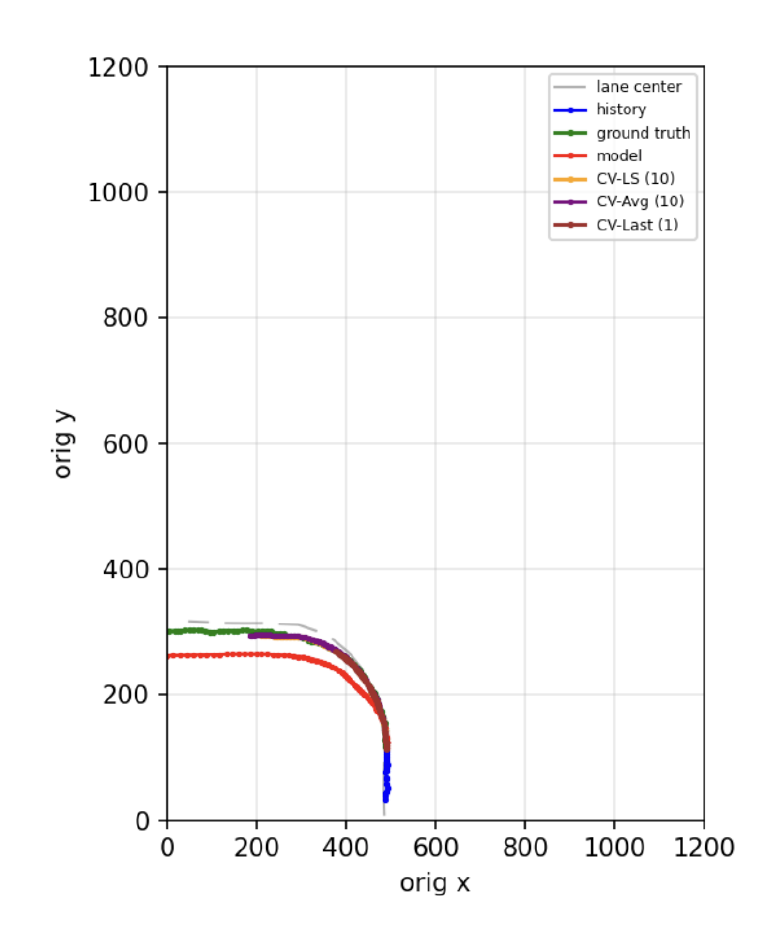}
    \caption{Trajectory comparison between enactor and constant velocity baseline. Prediction Horizon is 10 seconds.}
    \label{fig:traj-compare}
\end{figure}

\subsection{Results}
\label{sec:results_real}

Tables~\ref{tab:real_fde}--\ref{tab:real_ade} report Enactor's multi-horizon FDE and ADE on the held-out real-world segment, alongside the three constant-velocity baselines (CV last~$1$, CV avg~$10$, CV ls~$10$). The real-world section uses only constant-velocity baselines: the IntTrajSim baseline from the simulator regime was not trained or evaluated on real-world data. We report both Enactor-Smooth and Enactor-Raw because whether spline smoothing of the training input actually helps is itself the experimental question motivated in Section~\ref{sec:data_real}. Both variants are evaluated against the same held-out raw target trajectories; only the training-input preprocessing differs. Thus the comparison should be read as a training-preprocessing ablation, not as a comparison of online smoothing pipelines. Both beat every constant-velocity baseline at every horizon: Enactor-Smooth has FDE $0.76$, $1.43$, and $2.45$~m at $2$, $5$, and $10$~seconds; Enactor-Raw has FDE $0.83$, $1.28$, and $2.04$~m; the strongest constant-velocity baseline (CV ls~$10$) reaches $0.96$, $2.45$, and $3.61$~m. The gap is largest at longer horizons, where constant velocity cannot extrapolate curved trajectories, but Enactor's improvement is consistent across both variants and all three CV references. Among the constant-velocity variants, CV avg~$10$ has slightly lower ADE/FDE than CV last~$1$ but worse within-threshold rates: mean displacement and threshold success measure different parts of the error distribution, and a smoother velocity estimate can lower the mean while pushing more of the error mass just outside the success threshold.

\paragraph*{Smoothing trade-off across horizons}
Comparing Enactor-Smooth and Enactor-Raw in Tables~\ref{tab:real_fde} and \ref{tab:real_ade}, Enactor-Smooth has the edge at the $2$-second horizon (FDE@$2$~s: $0.76$ vs.\ $0.83$~m; ADE@$2$~s: $0.39$ vs.\ $0.44$~m). At longer horizons the two variants are close on average error: ADE is identical at $5$~s ($0.68$ vs.\ $0.68$~m) and within $10\%$ at $10$~s ($1.00$ vs.\ $1.09$~m). FDE separates more, in favor of Enactor-Raw ($1.28$ vs.\ $1.43$~m at $5$~s, $2.04$ vs.\ $2.45$~m at $10$~s), and within-threshold success rates (Table~\ref{tab:real_success}) favor Enactor-Raw at every horizon. The pattern is consistent with a low-pass filter on the input: filtering removes centroid jitter the model should not have to fit, which helps at short horizons, while the small bias smoothing introduces in the estimated current state matters more as the prediction window lengthens. We use Enactor-Smooth as the default at $2$~seconds and Enactor-Raw as the default at $5$ and $10$~seconds, with the caveat that the long-horizon gap is modest on average error. We treat displacement metrics (Tables~\ref{tab:real_fde}, \ref{tab:real_ade}) and within-threshold success rates (Table~\ref{tab:real_success}) as the primary real-world measures.

\paragraph*{Within-threshold rates}
Table~\ref{tab:real_success} reports within-threshold success rates for both variants. Enactor-Smooth reaches $77.42\%$, $84.74\%$, and $88.49\%$ within $1$~m at $2$~s, $2$~m at $5$~s, and $4$~m at $10$~s respectively; Enactor-Raw reaches $78.04\%$, $87.14\%$, and $92.98\%$ at the same thresholds, again favoring Enactor-Raw at the longer horizons. Both Enactor variants substantially exceed the three constant-velocity baselines at every horizon (the strongest CV variant, ls~$10$, reaches $72.10\%$, $78.23\%$, $81.80\%$).

\begin{table}[t]
\centering
\scriptsize
\setlength{\tabcolsep}{3pt}
\renewcommand{\arraystretch}{1.15}
\begin{tabularx}{\columnwidth}{l *{3}{>{\centering\arraybackslash}X}}
    \rowcolor{black!30}
    \textbf{Model} &
    \textbf{ADE @ 2 s (m)} &
    \textbf{ADE @ 5 s (m)} &
    \textbf{ADE @ 10 s (m)} \\
    \rowcolor{gray!15}
    Constant-velocity (last 1) & 0.58 & 1.24 & 1.83 \\
    \rowcolor{gray!15}
    Constant-velocity (avg 10) & 0.55 & 1.23 & 1.79 \\
    \rowcolor{gray!15}
    Constant-velocity (ls 10) & 0.46 & 0.97 & 1.23 \\
    \rowcolor{gray!15}
    Enactor-Raw          & 0.44 & 0.68 & 1.00 \\
    \rowcolor{gray!15}
    Enactor-Smooth       & 0.39 & 0.68 & 1.09 \\
\end{tabularx}
\caption{Multi-horizon Average Displacement Error on real-world data (meters).}
\label{tab:real_ade}
\end{table}

\begin{table}[t]
\centering
\scriptsize
\setlength{\tabcolsep}{3pt}
\renewcommand{\arraystretch}{1.15}
\begin{tabularx}{\columnwidth}{l *{3}{>{\centering\arraybackslash}X}}
    \rowcolor{black!30}
    \textbf{Model} &
    \textbf{FDE @ 2 s (m)} &
    \textbf{FDE @ 5 s (m)} &
    \textbf{FDE @ 10 s (m)} \\
    \rowcolor{gray!15}
    Constant-velocity (last 1) & 1.16 & 2.92 & 4.70 \\
    \rowcolor{gray!15}
    Constant-velocity (avg 10) & 1.12 & 2.90 & 4.61 \\
    \rowcolor{gray!15}
    Constant-velocity (ls 10) & 0.96 & 2.45 & 3.61 \\
    \rowcolor{gray!15}
    Enactor-Raw          & 0.83 & 1.28 & 2.04 \\
    \rowcolor{gray!15}
    Enactor-Smooth       & 0.76 & 1.43 & 2.45 \\
\end{tabularx}
\caption{Multi-horizon Final Displacement Error on real-world data (meters).}
\label{tab:real_fde}
\end{table}

\begin{table}[t]
\centering
\scriptsize
\setlength{\tabcolsep}{3pt}
\renewcommand{\arraystretch}{1.15}
\begin{tabularx}{\columnwidth}{l *{3}{>{\centering\arraybackslash}X}}
    \rowcolor{black!30}
    \textbf{Model} &
    \textbf{Within 1\,m @ 2\,s} &
    \textbf{Within 2\,m @ 5\,s} &
    \textbf{Within 4\,m @ 10\,s} \\
    \rowcolor{gray!15}
    Constant velocity (last 1)   & 69.96\% & 75.88\% & 79.21\% \\
    \rowcolor{gray!15}
    Constant velocity (avg 10)   & 66.28\% & 70.31\% & 73.71\% \\
    \rowcolor{gray!15}
    Constant velocity (ls 10)   & 72.10\% & 78.23\% & 81.80\% \\
    \rowcolor{gray!15}
    Enactor-Raw         & 78.04\% & 87.14\% & 92.98\% \\
    \rowcolor{gray!15}
    Enactor-Smooth      & 77.42\% & 84.74\% & 88.49\% \\
\end{tabularx}
\caption{Within-threshold success rates at multiple horizons on real-world data. Each percentage is the fraction of evaluation samples whose predicted position lies within the indicated threshold of ground truth. The denominator is the number of valid evaluation samples at the corresponding horizon ($N=60{,}472$ at $2$~s, $N=48{,}597$ at $5$~s, $N=41{,}802$ at $10$~s). Results are reported on vehicles only; pedestrians are out of the scope of this evaluation (see Section~\ref{sec:intro}).}
\label{tab:real_success}
\end{table}

\section{Conclusions}
\label{sec:conclusion}

We have presented Enactor, a transformer-based generative model for closed-loop microsimulation of signalized traffic intersections. In the simulator regime, the polar-coordinate state representation and the closed-loop training schedule give a model whose macroscopic speed and travel-time distributions match those of SUMO itself, with KL divergence well below that of a recent transformer-based baseline (over an order of magnitude lower on travel time at both intersections, and roughly $5\times$ lower on speed at Site~1). Enactor also improves on the same baseline on most trajectory-reconstruction metrics at both intersections (FDE at Site~1 and Site~2; ADE at Site~1, while ADE at Site~2 is comparable). On safety metrics, Enactor reduces red-light violations by more than an order of magnitude relative to IntTrajSim at both intersections.

The defining feature of the evaluation is that the actor cohort evolves over the rollout: actors enter on one approach, traverse the intersection, leave on another, and new actors arrive continuously at the perimeter. Sustaining a stable rollout of thousands of seconds against this non-stationary set, rather than against a fixed cohort at a fixed prompt time, is what distinguishes a microsimulator from a trajectory predictor, and is the setting Enactor is evaluated in. We have argued that stable long-horizon rollout, aggregate-flow realism, and intersection-aware safety realism are distinct objectives: the polar representation and closed-loop curriculum address the first two; the leader rear-bumper feature is the single change with the largest effect on the third, reducing red-light violations and TTC events relative to a base variant without it. Across the two intersection geometries, aggregate-flow performance is consistent and red-light compliance is preserved, while TTC remains a limitation. Close-range interaction realism is the main remaining limitation of the current model.

For the real-world setting, we have shown that the same architecture can be trained and evaluated on naturalistic trajectories from a fish-eye camera processed by a YOLO-based detection-and-tracking system, with appropriate preprocessing and a single-pass evaluation protocol that fits the recording. Both Enactor-Smooth and Enactor-Raw beat a constant-velocity baseline at every horizon evaluated. Enactor-Smooth has a small edge at $2$~seconds; at longer horizons the two variants are close on average error, with Enactor-Raw retaining a modest advantage on FDE and within-threshold success --- the behavior one would expect from a low-pass filter as the prediction horizon varies. Within-threshold success rates of $77$--$93\%$ at $1$--$4$~m thresholds across $2$--$10$~s horizons (Table~\ref{tab:real_success}) suggest the predictions are accurate enough for offline trajectory analysis and microsimulation-calibration workflows; online deployment would require a causal preprocessing pipeline and broader field validation. A qualitative comparison of predictions across a range of trajectory types is the next step.

The methodological lesson on the real-world side: upstream tracking noise affects the learned dynamics. Smoothing can suppress short-horizon jitter but may introduce bias that matters at longer horizons. Future work will target close-range interaction realism (the TTC regime) directly, through architectural changes that address it head-on; extend the real-world deployment to multiple cameras and longer recordings, so we can evaluate against rare and safety-critical events that a single hour of data does not cover; explore transfer-learning strategies that adapt weights trained at one intersection to new geometries without full retraining; and extend the actor-centric formulation to treat pedestrian behavior as a first-class target rather than context.


\bibliographystyle{IEEEtran}
\bibliography{paper}

@INPROCEEDINGS{intersectionnet,
  author={Wu, Aotian and Ranjan, Yash and Sengupta, Rahul and Rangarajan, Anand and Ranka, Sanjay},
  booktitle={2024 IEEE Intelligent Vehicles Symposium (IV)}, 
  title={A Data-driven Approach for Probabilistic Traffic Prediction and Simulation at Signalized Intersections}, 
  year={2024},
  volume={},
  number={},
  pages={3092-3099},
  doi={10.1109/IV55156.2024.10588424}}

@inproceedings{metrics,
author = {Ranjan, Yash and Sengupta, Rahul and Rangarajan, Anand and Ranka, Sanjay},
title = {Evaluating Generative Vehicle Trajectory Models for nbsp;Traffic Intersection Dynamics},
year = {2025},
isbn = {978-981-96-8294-2},
publisher = {Springer-Verlag},
address = {Berlin, Heidelberg},
url = {https://doi.org/10.1007/978-981-96-8295-9_19},
doi = {10.1007/978-981-96-8295-9_19},
booktitle = {Data Science: Foundations and Applications: 29th Pacific-Asia Conference on Knowledge Discovery and Data Mining, PAKDD 2025, Sydney, NSW, Australia, June 10-13, 2025, Proceedings, Part VI},
pages = {262–274},
numpages = {13},
location = {Sydney, NSW, Australia}
}

@inproceedings{trafficbots2023,
  title={TrafficBots: Towards World Models for Autonomous Driving Simulation and Motion Prediction},
  author={Zhang, L. and Gao, J. and Li, W. and others},
  booktitle={IEEE Intl. Conf. on Robotics and Automation (ICRA)},
  year={2023},
  doi={10.1109/ICRA48891.2023.10160880}
}

@inproceedings{transworldng2023,
  title={TransWorldNG: Traffic Simulation via Foundation Model},
  author={Zhang, Y. and Chen, X. and Liu, Y. and others},
  booktitle={IEEE Intl. Conf. on Smart Computing (SMARTCOMP)},
  year={2023},
  doi={10.1109/SMARTCOMP58114.2023.00076}
}

@article{beyondsimulation2025,
  title={Beyond Simulation: Benchmarking World Models for Planning and Causality in Autonomous Driving},
  author={Zheng, K. and Gao, J. and Li, W. and others},
  journal={arXiv preprint arXiv:2508.01922},
  year={2025}
}

@inproceedings{kigan2024,
  title={KI-GAN: Knowledge-Informed Generative Adversarial Networks for Vehicle Trajectory Prediction at Signalized Intersections},
  author={Wei, C. and Zhang, L. and Li, W. and others},
  booktitle={IEEE/CVF Conf. on Computer Vision and Pattern Recognition Workshops (CVPRW)},
  year={2024},
  doi={10.1109/CVPRW59438.2024.01117}
}

@article{hdgt2023,
  title={HDGT: Heterogeneous Driving Graph Transformer for Multi-Agent Trajectory Prediction via Scene Encoding},
  author={Jia, X. and Wu, P. and Chen, L. and others},
  journal={IEEE Trans. on Pattern Analysis and Machine Intelligence (TPAMI)},
  year={2023},
  doi={10.1109/TPAMI.2023.3298301}
}

@misc{kuncheria2025,
      title={Beyond Centrality: Understanding Urban Street Network Typologies Through Intersection Patterns}, 
      author={Anu Kuncheria and Joan L. Walker and Jane Macfarlane},
      year={2025},
      eprint={2511.06747},
      archivePrefix={arXiv},
      primaryClass={cs.SI},
      url={https://arxiv.org/abs/2511.06747}, 
}

@misc{montali2023,
      title={The Waymo Open Sim Agents Challenge}, 
      author={Nico Montali and John Lambert and Paul Mougin and Alex Kuefler and Nick Rhinehart and Michelle Li and Cole Gulino and Tristan Emrich and Zoey Yang and Shimon Whiteson and Brandyn White and Dragomir Anguelov},
      year={2023},
      eprint={2305.12032},
      archivePrefix={arXiv},
      primaryClass={cs.CV},
      url={https://arxiv.org/abs/2305.12032}, 
}

@misc{zhang2025,
      title={Relative Position Matters: Trajectory Prediction and Planning with Polar Representation}, 
      author={Bozhou Zhang and Nan Song and Bingzhao Gao and Li Zhang},
      year={2025},
      eprint={2508.11492},
      archivePrefix={arXiv},
      primaryClass={cs.RO},
      url={https://arxiv.org/abs/2508.11492}, 
}

@misc{gao2020,
      title={VectorNet: Encoding HD Maps and Agent Dynamics from Vectorized Representation}, 
      author={Jiyang Gao and Chen Sun and Hang Zhao and Yi Shen and Dragomir Anguelov and Congcong Li and Cordelia Schmid},
      year={2020},
      eprint={2005.04259},
      archivePrefix={arXiv},
      primaryClass={cs.CV},
      url={https://arxiv.org/abs/2005.04259}, 
}

@inproceedings{vaswani2017,
      title={Attention Is All You Need},
      author={Vaswani, Ashish and Shazeer, Noam and Parmar, Niki and Uszkoreit, Jakob and Jones, Llion and Gomez, Aidan N. and Kaiser, Lukasz and Polosukhin, Illia},
      booktitle={Advances in Neural Information Processing Systems (NeurIPS)},
      year={2017},
}

@misc{shi2023mtr,
      title={Motion Transformer with Global Intention Localization and Local Movement Refinement}, 
      author={Shaoshuai Shi and Li Jiang and Dengxin Dai and Bernt Schiele},
      year={2023},
      eprint={2209.13508},
      archivePrefix={arXiv},
      primaryClass={cs.CV},
      url={https://arxiv.org/abs/2209.13508}, 
}

@conference{enactor_vehits,
author={Yash Ranjan and Rahul Sengupta and Anand Rangarajan and Sanjay Ranka},
title={Enactor: From Traffic Simulators to Surrogate World Models},
booktitle={Proceedings of the 12th International Conference on Vehicle Technology and Intelligent Transport Systems - VEHITS},
year={2026},
pages={430-438},
publisher={SciTePress},
organization={INSTICC},
doi={10.5220/0014837100004030},
isbn={978-989-758-831-0},
issn={2184-495X},
}

@misc{ranjan2025,
      title={IntTrajSim: Trajectory Prediction for Simulating Multi-Vehicle driving at Signalized Intersections}, 
      author={Yash Ranjan and Rahul Sengupta and Anand Rangarajan and Sanjay Ranka},
      year={2025},
      eprint={2506.08957},
      archivePrefix={arXiv},
      primaryClass={cs.AI},
      url={https://arxiv.org/abs/2506.08957}, 
}

@article{NI2020102137,
title = {Limitations of current traffic models and strategies to address them},
journal = {Simulation Modelling Practice and Theory},
volume = {104},
pages = {102137},
year = {2020},
issn = {1569-190X},
doi = {https://doi.org/10.1016/j.simpat.2020.102137},
url = {https://www.sciencedirect.com/science/article/pii/S1569190X20300769},
author = {Daiheng Ni}
}

@inproceedings{ettinger2021large,
  title={Large Scale Interactive Motion Forecasting for Autonomous Driving: The Waymo Open Motion Dataset},
  author={Ettinger, Scott and Timofeev, Aleksei and Krivokon, Maxim and Gao, Amy and Joshi, Aditya and Zhang, Yu and Shlens, Jonathon and Anguelov, Dragomir},
  booktitle={Proceedings of the IEEE/CVF International Conference on Computer Vision (ICCV)},
  pages={4378-4387},
  year={2021}
}

@misc{wang2023buildingtransportationfoundationmodel,
      title={Building Transportation Foundation Model via Generative Graph Transformer}, 
      author={Xuhong Wang and Ding Wang and Liang Chen and Yilun Lin},
      year={2023},
      eprint={2305.14826},
      archivePrefix={arXiv},
      primaryClass={cs.LG},
      url={https://arxiv.org/abs/2305.14826}, 
}

@misc{tan2025scenediffusercityscaletrafficsimulation,
      title={SceneDiffuser++: City-Scale Traffic Simulation via a Generative World Model}, 
      author={Shuhan Tan and John Lambert and Hong Jeon and Sakshum Kulshrestha and Yijing Bai and Jing Luo and Dragomir Anguelov and Mingxing Tan and Chiyu Max Jiang},
      year={2025},
      eprint={2506.21976},
      archivePrefix={arXiv},
      primaryClass={cs.LG},
      url={https://arxiv.org/abs/2506.21976}, 
}

@misc{salzmann2021,
      title={Trajectron++: Dynamically-Feasible Trajectory Forecasting With Heterogeneous Data}, 
      author={Tim Salzmann and Boris Ivanovic and Punarjay Chakravarty and Marco Pavone},
      year={2021},
      eprint={2001.03093},
      archivePrefix={arXiv},
      primaryClass={cs.RO},
      url={https://arxiv.org/abs/2001.03093}, 
}

@article{chen2019bayesian,
  author={Chen, Xiaoxia and Zhou, Lei and Li, Lin},
  title={Bayesian network for red-light-running prediction at signalized intersections},
  journal={Journal of Intelligent Transportation Systems},
  volume={23},
  number={2},
  pages={120--132},
  year={2019},
  month=mar,
}

@article{yang2023hierarchical,
  author={Yang, Zhenyu and Zhang, Ruochen and Pandey, Gaurav and Masoud, Neda and Liu, Henry X.},
  title={A hierarchical vehicle behavior prediction framework with traffic signals and interactive agents},
  journal={IEEE Transactions on Intelligent Transportation Systems},
  volume={24},
  number={10},
  pages={11066--11079},
  year={2023},
  month=oct,
}

@article{trentin2023multimodal,
  author={Trentin, Vinicius and Artu{\~n}edo, Antonio and Godoy, Jorge and Villagra, Jorge},
  title={Multi-modal interaction-aware motion prediction at unsignalized intersections},
  journal={IEEE Transactions on Intelligent Vehicles},
  volume={8},
  number={5},
  pages={3349--3365},
  year={2023},
  month=may,
}

@article{sun2023mmhsta,
  author={Sun, Yang and Xu, Tao and Li, Jian and Chu, Yan and Ji, Xianglin},
  title={{MMH-STA}: A macro-micro-hierarchical spatio-temporal attention method for multi-agent trajectory prediction in unsignalized roundabouts},
  journal={IEEE Transactions on Vehicular Technology},
  volume={72},
  number={9},
  pages={11237--11250},
  year={2023},
  month=sep,
}

@inproceedings{zhang2022d2tpred,
  author={Zhang, Yi and Wang, Wenjun and Guo, Wei and Lv, Pei and Xu, Mingliang and Chen, Wei and Manocha, Dinesh},
  title={{D2-TPred}: Discontinuous dependency for trajectory prediction under traffic lights},
  booktitle={European Conference on Computer Vision (ECCV)},
  pages={522--539},
  year={2022},
}

@misc{oh2019impact,
  author={Oh, Geunseob and Peng, Huei},
  title={Impact of traffic lights on trajectory forecasting of human-driven vehicles near signalized intersections},
  year={2019},
  eprint={1909.04826},
  archivePrefix={arXiv},
}

@inproceedings{roy2019gan,
  author={Roy, Debaditya and Ishizaka, Tetsuhiro and Mohan, C. Krishna and Fukuda, Atsushi},
  title={Vehicle trajectory prediction at intersections using interaction based generative adversarial networks},
  booktitle={IEEE Intelligent Transportation Systems Conference (ITSC)},
  pages={2318--2323},
  year={2019},
  address={Auckland, New Zealand},
}

@article{zhang2021turning,
  author={Zhang, Tao and Song, Wenjie and Fu, Mengyin and Yang, Yi and Wang, Meiling},
  title={Vehicle motion prediction at intersections based on the turning intention and prior trajectories model},
  journal={IEEE/CAA Journal of Automatica Sinica},
  volume={8},
  number={10},
  pages={1657--1666},
  year={2021},
  month=oct,
}

@article{geng2023transferable,
  author={Geng, Maosi and Li, Junyi and Li, Cong and Xie, Ning and Chen, Xiqun and Lee, Der-Horng},
  title={Adaptive and simultaneous trajectory prediction for heterogeneous agents via transferable hierarchical transformer network},
  journal={IEEE Transactions on Intelligent Transportation Systems},
  volume={24},
  number={10},
  pages={11479--11492},
  year={2023},
  month=oct,
}

@article{zhang2023aitp,
  author={Zhang, Kunpeng and Zhao, Liang and Dong, Chuanyou and Wu, Lan and Zheng, Lihua},
  title={{AI-TP}: Attention-based interaction-aware trajectory prediction for autonomous driving},
  journal={IEEE Transactions on Intelligent Vehicles},
  volume={8},
  number={1},
  pages={73--83},
  year={2023},
  month=jan,
}

@article{messaoud2021attention,
  author={Messaoud, Kaouther and Yahiaoui, Itheri and Verroust-Blondet, Anne and Nashashibi, Fawzi},
  title={Attention based vehicle trajectory prediction},
  journal={IEEE Transactions on Intelligent Vehicles},
  volume={6},
  number={1},
  pages={175--185},
  year={2021},
  month=mar,
}

@article{chen2022intentionaware,
  author={Chen, Xiaobo and Zhang, Huanjia and Zhao, Feng and Hu, Yu and Tan, Chenyi and Yang, Jian},
  title={Intention-aware vehicle trajectory prediction based on spatial-temporal dynamic attention network for internet of vehicles},
  journal={IEEE Transactions on Intelligent Transportation Systems},
  volume={23},
  number={10},
  pages={19471--19483},
  year={2022},
  month=oct,
}

@inproceedings{alahi2016sociallstm,
  author={Alahi, Alexandre and Goel, Kratarth and Ramanathan, Vignesh and Robicquet, Alexandre and Fei-Fei, Li and Savarese, Silvio},
  title={Social {LSTM}: Human trajectory prediction in crowded spaces},
  booktitle={IEEE Conference on Computer Vision and Pattern Recognition (CVPR)},
  pages={961--971},
  year={2016},
}

@inproceedings{deo2018convolutional,
  author={Deo, Nachiket and Trivedi, Mohan M.},
  title={Convolutional social pooling for vehicle trajectory prediction},
  booktitle={IEEE Conference on Computer Vision and Pattern Recognition Workshops (CVPRW)},
  pages={1468--1476},
  year={2018},
}

@inproceedings{gupta2018socialgan,
  author={Gupta, Agrim and Johnson, Justin and Fei-Fei, Li and Savarese, Silvio and Alahi, Alexandre},
  title={Social {GAN}: Socially acceptable trajectories with generative adversarial networks},
  booktitle={IEEE Conference on Computer Vision and Pattern Recognition (CVPR)},
  pages={2255--2264},
  year={2018},
}

@Article{jimaging8040101,
AUTHOR = {Huang, Xiaohui and He, Pan and Rangarajan, Anand and Ranka, Sanjay},
TITLE = {Machine-Learning-Based Real-Time Multi-Camera Vehicle Tracking and Travel-Time Estimation},
JOURNAL = {Journal of Imaging},
VOLUME = {8},
YEAR = {2022},
NUMBER = {4},
ARTICLE-NUMBER = {101},
URL = {https://www.mdpi.com/2313-433X/8/4/101},
PubMedID = {35448228},
ISSN = {2313-433X},
DOI = {10.3390/jimaging8040101}
}

\begin{figure*}[t]
    \centering
    \includegraphics[height=0.8\textheight]{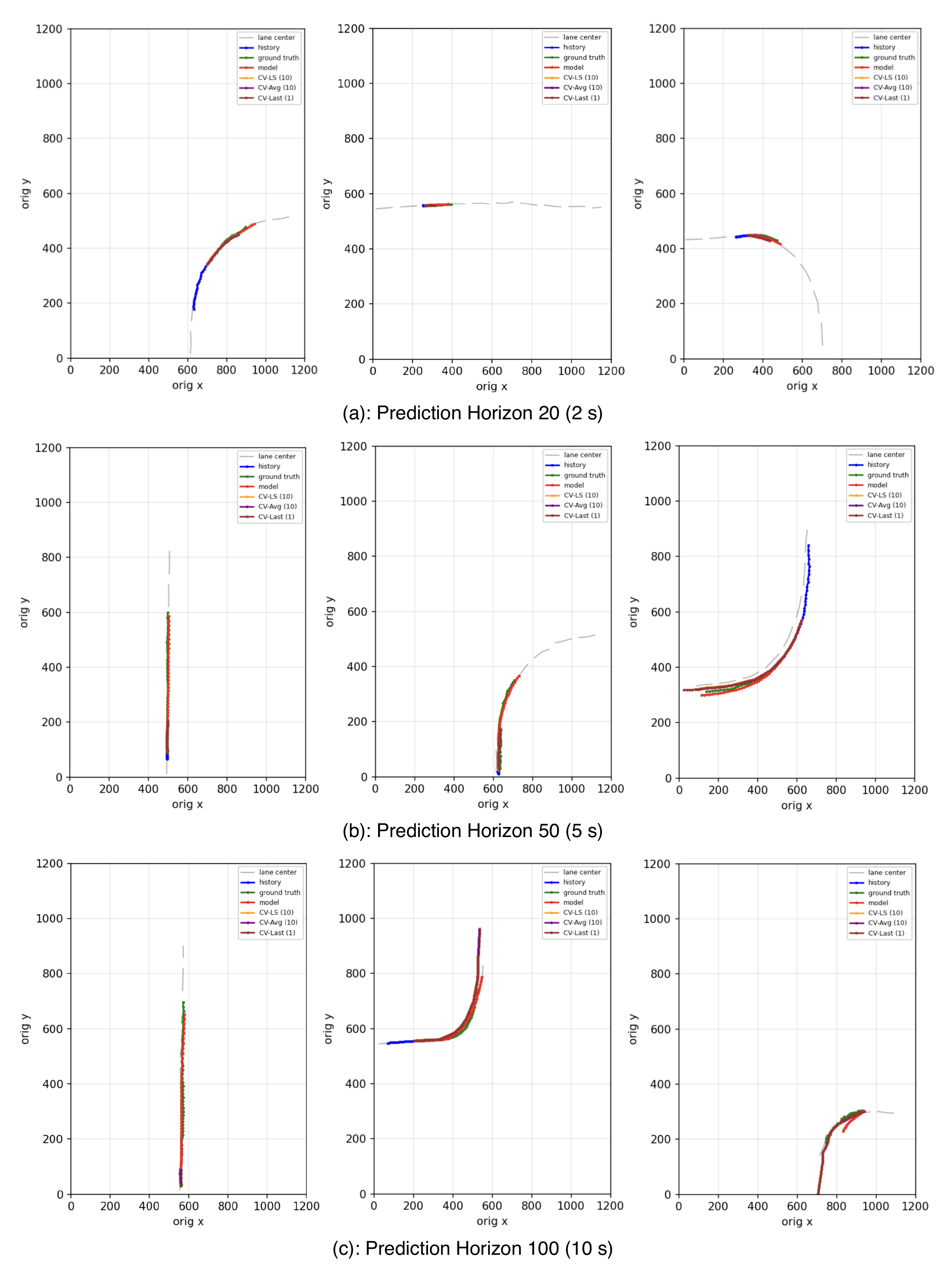}
    \caption{Trajectory comparison between enactor and constant velocity baselines with different prediction horizons.}
    \label{fig:traj-all}
\end{figure*}

\end{document}